\documentclass[twoside,leqno,twocolumn]{article}
 \pdfoutput=1
\usepackage[letterpaper]{geometry}

\usepackage{ltexpprt}
\usepackage{hyperref}

\usepackage{placeins}
\usepackage{xcolor} 
\usepackage{amsmath}
\usepackage{amsfonts}
\usepackage{graphicx}
\usepackage{float}
\usepackage{tikz}
\usetikzlibrary{arrows.meta,arrows}
\usetikzlibrary{shapes.geometric}

\usepackage[square,numbers]{natbib}
\usepackage{natbibspacing}
\definecolor{cornellred}{rgb}{0.7, 0.11, 0.11}

\definecolor{steelblue}{rgb}{0.2745, 0.5098, 0.7059}

\begin{document}

\newcommand\relatedversion{}
\renewcommand\relatedversion{\thanks{The full version of the paper can be accessed at \protect\url{https://arxiv.org/abs/1902.09310}}} 

\title{IB-GAN: A Unified Approach for Multivariate Time Series Classification under Class Imbalance}

\author{%
  Grace Deng \\
  Dept. of Statistics \& Data Science\\
  Cornell University\\
  \texttt{gd345@cornell.edu} \\
   \and
   Cuize Han \\
   Amazon Search \\
   \texttt{cuize@amazon.com} \\
  \and
  Tommaso Dreossi \\
   Amazon Search \\
  \texttt{tommasodreossi@gmail.com}
    \and
   Clarence Lee\\
   Cornell University\\
   \texttt{clarence.lee@cornell.edu} \\
  \and
   David S.\ Matteson\\
   Dept. of Statistics \& Data Science \\
   Cornell University\\
   \texttt{dm484@cornell.edu} \\
}

\date{}

\maketitle


\fancyfoot[R]{\scriptsize{
Unauthorized reproduction of this article is prohibited}}





\begin{abstract} 
Classification of large multivariate time series with strong class imbalance is an important task in real-world applications. Standard methods of class weights, oversampling, or parametric data augmentation do not always yield significant improvements for predicting minority classes of interest. Non-parametric data augmentation with Generative Adversarial Networks (GANs) offers a promising solution. We propose Imputation Balanced GAN (IB-GAN), a novel method that joins data augmentation and classification in a one-step process via an imputation-balancing approach. IB-GAN uses imputation and resampling techniques to generate higher quality samples from randomly masked vectors than from white noise, and augments classification through a class-balanced set of real and synthetic samples. Imputation hyperparameter $p_{miss}$ allows for regularization of classifier variability by tuning innovations introduced via generator imputation. IB-GAN is simple to train and model-agnostic, pairing any deep learning classifier with a generator-discriminator duo and resulting in higher accuracy for under-observed classes. Empirical experiments on open-source UCR data and proprietary 90K product dataset show significant performance gains against state-of-the-art parametric and GAN baselines.

\end{abstract}


\section{Background and Motivation}
Multivariate time series classification (MTSC) is a growing field where complex features are collected in the form of time-indexed sequences with varying lengths. Often a single observation unit (product, video, recommendation, etc.) can be described by multiple time series metrics with strong inter-temporal dependencies and a mixture of numeric, categorical, and semantic features. Many recent works using variants of CNNs or RNNs have shown impressive performance in MTSC tasks \citep{fawaz2019deep, karim2017lstm, zhao2017convolutional, zheng2016exploiting, karim2019multivariate}. 
However, many real-world datasets consists of strong class imbalance where far fewer samples are observed from minority classes of interest. The three baseline methods for addressing imbalances are class weights, upsampling, and downsampling, each with drawbacks \cite{buda2018systematic, Weiss2007CostSensitiveLV}. Downsampling leads to poor classifier performance when too many samples are discarded, while upsampling leads to overfitting by reusing data from the minority class. Data augmentation techniques such as SMOTE \cite{chawla2002smote, dablain2021deepsmote} propose creating additional synthetic samples from oversampling and linear interpolation of k-nearest neighbors, but is still a parametric model constrained by computation time and does not generalize well to high-dimensional datasets \cite{lusa2012evaluation}. Techniques for image augmentation \cite{shorten2019survey} include transformations, cropping, noise injection, and random erasing \cite{zhong2020random}; however, far fewer methods have been developed for time series or more general data types.

Large-scale non-parametric data augmentation~\cite{fawaz2018data, shorten2019survey, mariani2018bagan} with Generative Adversarial Networks (GANs) \cite{goodfellow2014generative} poses a promising solution. Our paper presents Imputation Balanced GAN (IB-GAN), a novel imputation balancing approach to data augmentation that can pair any deep learning classifier with a generator-discriminator model to address class imbalance in MTSC.

\subsection{Related Works.}

Modern MTSC methods involve combinations of Convolutional Neural Networks (CNNs), Recurrent Neural Nets (RNNs), and Long Short-Term Memory (LSTM) networks, with variants such as stacked CNNs and Squeeze-and-Excitation block \cite{karim2019multivariate}, deep CNNs \cite{zheng2016exploiting}, and attention networks \cite{zhang2020tapnet}. These works are not designed for imbalanced data. A 2019 survey paper \cite{fawaz2019deep} states that very little has been done for imbalanced classes in time series classification other than class weights. Fawaz et al \cite{fawaz2018data} and Tran et al \cite{tran2017bayesian} showed data augmentation with deep learning models improves image classification. We propose that any of these classifiers with proven track record for MTSC can be paired with GANs to augment the training process, i.e., include synthetic data for under-sampled classes.


The GAN has two competing models: the \textit{generator} creates synthetic data by learning the true data distribution, and the \textit{discriminator} distinguishes real and synthetic samples. The vanilla GAN \cite{goodfellow2014generative} has uses in many generative tasks \cite{oza2019progressive, sheng2019unsupervised}.  Deep-Convolutional GANs (DC-GAN) use convolution layers \cite{radford2015unsupervised} to improve training stability via learning feature maps and object representations. Conditional GANs \cite{mirza2014conditional} extends DC-GAN and conditions on auxiliary information such as class labels. Additional techniques such as batch normalization\cite{salimans2016improved}, leaky-relu activation, max-pooling, and dropout all can improve GAN training. GANs specifically designed for imputation such as GAIN \cite{yoon2018gain}, Colla-GAN \cite{lee2019collagan}, Mis-GAN \cite{li2019misgan,  luo2018multivariate}, and for sequence generation \cite{yoon2019time, esteban2017real, zhang2021missing, guo2019data} have also been introduced.

More recent works focus on generating better samples. The Balancing GAN (BAGAN) adopts an autoencoder-decoder structure to learn class conditioning in the latent space \cite{mariani2018bagan}. The Auxiliary Classifier GAN (AC-GAN) \cite{odena2017conditional, gong2019twin} proposed a secondary classifier embedded in the discriminator in order to stabilize GAN training, but does not apply it directly for classification; a second-stage classifier is required. The Rumi formulation for GANs \cite{NEURIPS2020_29405e2a} trains the generator to specifically produce samples from a desired class. The Triple-GAN \cite{li2017triple} uses a generator to produce pseudo-samples and classifier to produce pseudo-labels, and the discriminant tries to predict whether a psuedo-pair of (sample, label) is real or synthetic. All of these methods (BAGAN, AC-GAN, etc.) were designed for image data in mind. 

\begin{figure}[H]
    \centering
    \includegraphics[width=0.4\textwidth]{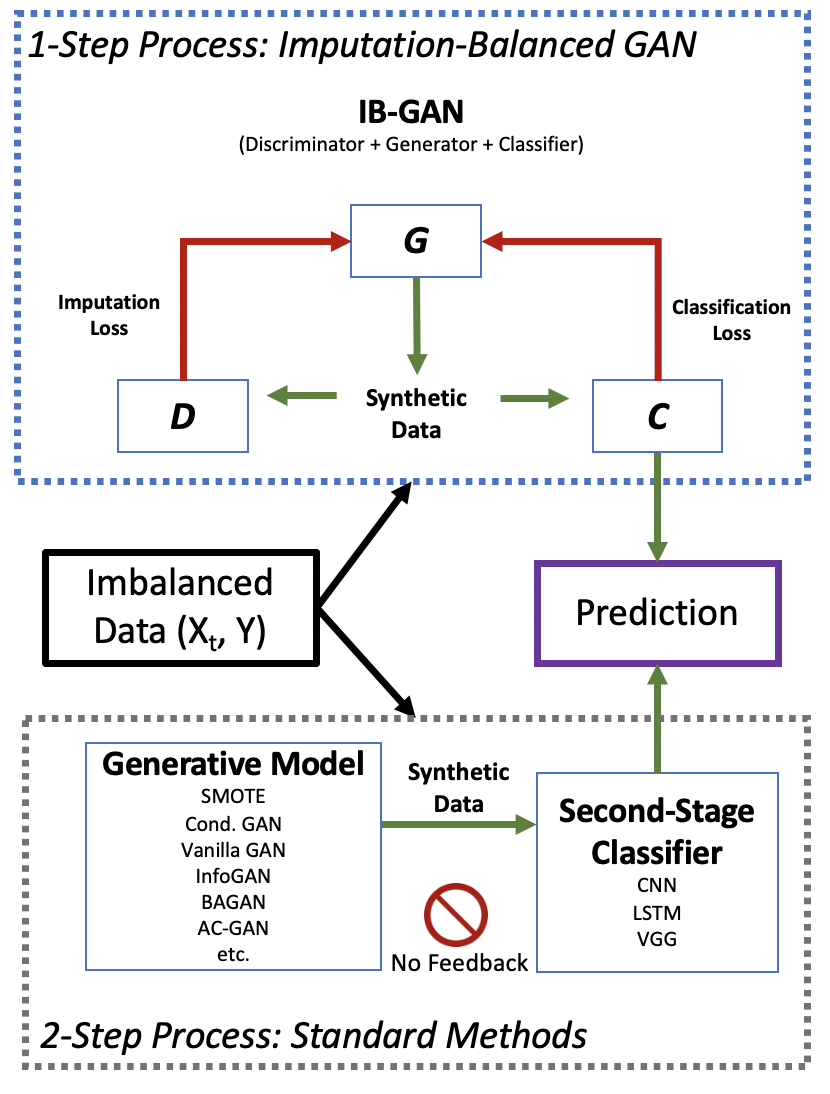}
    \caption{IB-GAN versus the two-step process for existing GAN data augmentation methods. IB-GAN directly utilizes $C$ model in $(D, G, C)$ triplet for classification; see Figure \ref{fig:flow} for detailed set-up.}
    \label{fig:one-stage}
\end{figure}

These GAN methods suffer from two drawbacks. First, they are specifically designed for data generation but requires a separate classification step, resulting in a two-stage training process where the generative model does not learn from classification feedback (see Figure \ref{fig:one-stage}). Second, these methods generate synthetic data from white or latent noise \cite{odena2017conditional, mariani2018bagan}, which risks unrealistic or repetitive samples due to mode collapse. The Imputation Balanced GAN (IB-GAN) will address these two challenges through a joint training process and a two-pronged imputation-balancing approach.

 

\vspace*{-\belowdisplayskip}

\subsection{Contributions.}
We propose the Imputation Balanced GAN (IB-GAN) for imbalanced classification with three key contributions:
\begin{enumerate} \parskip0pt  
    \item We present a unified one-step process (Figure \ref{fig:one-stage}) for joint data augmentation and classification under class imbalance, consisting of a triplet of generator $G$, discriminator $D$, and classifier $C$ models that is model agnostic to different neural net architectures. 
    \item We adapt a novel imputation approach for generating synthetic samples from randomly masked vectors; data quality is improved via direct feedback from classification and imputation losses. A tuning parameter $p_{miss}$ regulates innovations in generator imputations while preventing mode collapse common in standard GANs initialized by white noise.
    \item We balance classifier training via resampling techniques and synthetic samples for under-observed classes, with significant performance gains against state-of-the-art baselines.
\end{enumerate}

Full overview of IB-GAN is outlined in Figure \ref{fig:flow}. Theoretically, a range of GAN variants (vanilla \cite{goodfellow2014generative}, conditional \cite{mirza2014conditional}, Info-GAN \cite{chen2016infogan} etc.) and deep learning classifiers (CNN, RNN, LSTM, VGG-16) is possible for IB-GAN given the model-agnostic framework; we explore these options in Section \ref{subsec:mts_class}. Elaborating on contribution (2), IB-GAN generator initializes data generation with randomly masked data vectors using the MCAR missing mechanism \cite{mealli2015clarifying}, which leverages existing information to produce better quality samples. Resulting imputations are analogous to image perturbations in computer vision applications \cite{zheng2016improving, poursaeed2018generative}. Empowered by these new weighted resampling and data masking steps, the final IB-GAN classifier generalizes well to hold-out sets with strong class imbalance.

\section{Methods and Technical Solutions}
We briefly discuss the challenge to train unbiased classifiers from highly imbalanced data. Methods such as class weights may cause numerical instability \cite{gong2016novel} and data augmentation, e.g., via GANs, poses a better solution. We then introduce the main IB-GAN framework and objective functions.

\subsection{Data Augmentation for Imbalanced Classification.} \label{subsec: imbalanced_classification}
Without loss of generality, we assume all random variables involved are discrete with probability measure $P$. Suppose the observed data is coming from the realization of random pair: $(\boldsymbol{X},Y)$. $Y$ is the class label which is from a finite class label set ${\cal Y}$. Denote the prior label probability
as $w_{y}=P(Y=y)$ for $y\in{\cal Y}$. $\boldsymbol{X}$ is a $k$
dimensional random vector (some characteristic features for class
$Y$), $\boldsymbol{X}=(X_{1},..,X_{k})$ where $X_{i}\in{\cal X}$.
Given $Y=y$, denote the conditional probability $P(\boldsymbol{X}=\boldsymbol{x}|Y=y)$
as $p_{y}(\boldsymbol{x})$. The task is to train a classifier $C$
that maps ${\cal X}^{k}$ to ${\cal Y}$ or equivalently predicts
the class label probability. We will use the latter definition of
the task. That is, for $\boldsymbol{x}\in{\cal X}^{k}$, $C(\boldsymbol{x})=\{c_{y}(\boldsymbol{x})\}_{y\in{\cal Y}}$,
where $0<=c_{y}(\boldsymbol{x})<=1$ and is the predicted probability that $\boldsymbol{x}$ belongs to class $y$. We have $\sum_{y\in{\cal Y}}c_{y}(\boldsymbol{x})=1,\forall\boldsymbol{x}\in{\cal X}^{k}$.
The classifier maximizing equal class weight negative cross-entropy loss: 
\vspace*{-\abovedisplayskip}
\begin{equation}
\begin{aligned}L_{class}(C) & = \sum_{y\in{\cal Y}} \mathbb{E}
\left[1\{Y=y\}\log c_{y}(\boldsymbol{X})\right] \\ &= \sum_{y\in{\cal Y},\boldsymbol{x}\in{\cal X}^{k}}w_{y}p_{y}(\boldsymbol{x})\log c_{y}(\boldsymbol{x})
\end{aligned}
\end{equation}

 is known as the Bayes classifier and can be expressed as $C^{*}(x)=\{c_{y}^{*}(\boldsymbol{x})\}_{y\in{\cal Y}}$,
where $c_{y}^{*}(\boldsymbol{x})=\frac{w_{y}p_{y}(\boldsymbol{x})}{\sum_{y'\in{\cal Y}}w_{y'}p_{y'}(\boldsymbol{x})}=\frac{P(Y=y,\boldsymbol{X}=\boldsymbol{x})}{P(\boldsymbol{X}=\boldsymbol{x})}=P(Y=y|\boldsymbol{X}=\boldsymbol{x})$.
This is the optimal classifier for the data distribution $(\boldsymbol{X},Y)$.
However, our focus is on minimizing overall classification error when
we treat each class equally. Then under this criterion, the Bayes
classifier is suboptimal due to highly unbalanced prior label probability
$w_{y}$ and is biased towards labels with high prior probability.
Let $U$ be a uniform random variable over the class label set ${\cal Y}$.
The joint distribution of $(\boldsymbol{X},U)$ is defined through
letting $P(\boldsymbol{X}=\boldsymbol{x}|U=y)\equiv p_{y}(\boldsymbol{x})$.
Then the optimal balanced class classifier is $\bar{C}^{*}(x)=\{\bar{c}_{y}^{*}(\boldsymbol{x})\}_{y\in{\cal Y}}$,
where $\bar{c}_{y}^{*}(\boldsymbol{x})=P(U=y|\boldsymbol{X}=\boldsymbol{x})=\frac{p_{y}(\boldsymbol{x})}{\sum_{y'\in{\cal Y}}p_{y'}(\boldsymbol{x})}$.

In order to derive the optimal balanced class classifier for data distribution $(X,Y)$ with class imbalance, a common method is inverse class weights in the loss. The classifier that maximizes the negative cross-entropy loss with inverse class weights:
\begin{equation}
\begin{aligned}L_{class}^{w}(C) &= \sum_{y\in{\cal Y}}w_{y}^{-1}\mathbb{E}\left[1\{Y=y\}\log c_{y}(\boldsymbol{X})\right]
\\ &= \sum_{y\in{\cal Y},\boldsymbol{x}\in{\cal X}^{k}}p_{y}(\boldsymbol{x})\log c_{y}(\boldsymbol{x})
\end{aligned}
\end{equation}
is $\bar{C}^{*}(x)$. Solving this empirical optimization problem, e.g., with with highly unbalanced inverse weights $w_{y}^{-1}$, can cause numerical instability issues.

A better solution is in the form of data augmentation. Suppose
we have additional samples for classifier training that
are realizations of random pair $(\boldsymbol{X}',Y')$. For example, additional training samples can be derived from resampling of existing data, transformations and rotations, or model based data augmentation. Let the prior label probability of the additional data distribution to be $w'_{y}=P(Y'=y)$ for $y\in{\cal Y}$ and the conditional probability $p'_{y}(\boldsymbol{x}):=P(\boldsymbol{X}'=\boldsymbol{x}|Y'=y)$.
We define the loss that combines the two source
of data through a hyperparameter $0<\alpha<1$ as 

\begin{equation}
\resizebox{0.5\textwidth}{!}{%
$\begin{aligned}L_{class}^{\alpha}(C) &= \alpha\sum_{y\in{\cal Y}}\mathbb{E}\left[1\{Y=y\}\log c_{y}(\boldsymbol{X})\right] \\ &+(1-\alpha)\sum_{y\in{\cal Y}}\mathbb{E}\left[1\{Y'=y\}\log c_{y}(\boldsymbol{X}')\right]\\
& = \sum_{y\in{\cal Y},\boldsymbol{x}\in{\cal X}^{k}}\left(\alpha w_{y}p_{y}(\boldsymbol{x})+(1-\alpha)w'_{y}p'_{y}(\boldsymbol{x})\right)\log c_{y}(\boldsymbol{x})
\end{aligned}$
}
\end{equation}

From Jensen's inequality, the optimal classifier that maximizes $L_{class}^{\alpha}(C)$
is $\tilde{C}^{*}(\boldsymbol{x})=\{\tilde{c}_{y}^{*}(\boldsymbol{x})\}_{y\in{\cal Y}}$,
where $\tilde{c}_{y}^{*}(\boldsymbol{x})=\frac{\alpha w_{y}p_{y}(\boldsymbol{x})+(1-\alpha)w'_{y}p'_{y}(\boldsymbol{x})}{\sum_{y'\in{\cal Y}}\left(\alpha w_{y'}p_{y'}(\boldsymbol{x})+(1-\alpha)w'_{y'}p'_{y'}(\boldsymbol{x})\right)}$.
Data augmentation can control over the prior label
probability $w_{y}'$. If $w_{y}'=\frac{|{\cal Y}|^{-1}-\alpha w_{y}}{1-\alpha}$
, $y\in{\cal Y}$ and we choose any $\alpha$ such that $\alpha<\frac{|{\cal Y}|^{-1}}{\max_{y}w_{y}}$,
then under optimal conditions where the augmented data
has the same conditional distribution as the original data,
$p'_{y}(\boldsymbol{x})\equiv p{}_{y}(\boldsymbol{x})$, we have $\tilde{C}^{*}(x)\equiv\bar{C}^{*}(x)$.
Thus with suitable choices of $w_{y}'$
and hyperparameter $\alpha,$ we can effectively train the balanced class classifier through combination of true and GAN-generated samples. We now introduce the IB-GAN which achieves data augmentation and balanced classification in an unified process.



\subsection{Imputation Balanced GAN.}
\begin{figure}[htp]
\begin{center}
\tikzstyle{block} = [rectangle, draw, font=\small,
    text width=6em, text centered, minimum height=2em, draw=steelblue, fill=steelblue!10]
\tikzstyle{block2} = [rectangle, draw, font=\small,
    text width=6em, text centered, minimum height=2em, draw=cornellred, fill=cornellred!10]
\tikzstyle{ioblock} = [trapezium, draw, trapezium left angle=-120, trapezium right angle=-60, fill=gray!10, font=\small,
    text width=3em, text centered, minimum height=2em]    
\tikzstyle{line} = [draw, -latex']
\begin{tikzpicture}[scale=0.5, node distance = 0.8cm, auto]
    \node [ioblock, text width=5em] (x_t) {Time series \\ $X_{t}$};
    \node [ioblock, text width=5em, right of=x_t, node distance=3.1cm] (x_fixed) {Metadata \\ $B$};
    \node [ioblock, right of=x_fixed, node distance=2.7cm] (labels) {Labels \\ $Y$};
    \node [block2, below of=x_fixed, node distance=1.5cm] (rebal) {Weighted Resample};
    
    \node [block2, below of=rebal, node distance=1.5cm] (mask) {Data Masking \\ $X_{mask}$};
    \node [block, below of=mask, node distance=1.5cm] (gen) {Generator \\ (Imputation)};
    
    \node [ioblock, text width=1.5em, right of=mask, node distance= 2.7cm, fill=cornellred!10, draw=cornellred] (p_miss){$p_{miss}$};
    
    \node [ioblock, below right of=gen, node distance=2.25cm, text width=6em] (synth) {Synthetic data \\ {$X'$}};
    \node [ioblock, below left of=gen, node distance=2.25cm, text width=5em] (real) {Real data \\ $X_{real}$};
    
    \node [block, below of=real, node distance=1.5cm] (disc) {Discriminator};
    \node [block, below of=synth, node distance=1.5cm] (class) {Classifier};
  
    \path [line] (x_t) -- (rebal);
    \path [line] (x_fixed) -- (rebal);
    \path [line] (labels) -- (rebal);
    
    \path [line, dashed] (p_miss) -- (mask);
    \path [line] (rebal) -- (mask);
    \path [line] (mask) -- (gen);
    \path [line] (gen) -- (synth);
    
    \path [line] (real) -- (disc);
    \path [line] (real) -- (class);
    \path [line] (synth) -- (disc);
    \path [line] (synth) -- (class);
    
    \path [line] (rebal) -| (real);

\end{tikzpicture}
\caption{IB-GAN triple-model framework (blue) includes novel resampling and data masking steps (red) for joint data augmentation and classification.}
\label{fig:flow}
\end{center}
\end{figure}
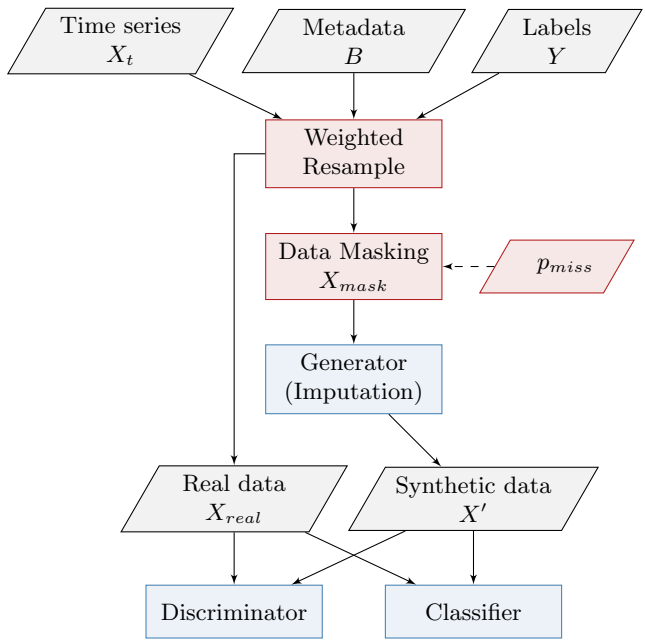

A standard GAN consists of two models \cite{goodfellow2014generative}, the generator $G$ that learns the underlying data distribution $p_{data}$ for data vector $X$ from noise distribution $Z$, and the discriminator $D$ that classifies whether the sample is from the true data distribution or a synthetic sample. For the IB-GAN (Figure \ref{fig:flow}), we introduce a new component $C$, the classifier that predicts which class $Y$ a sample belongs to, regardless of whether it is real or synthetic. Novel to prior work \cite{odena2017conditional}\cite{ tran2017bayesian}\cite{li2017triple}, the IB-GAN separates the discriminator and classifier into two different models with separate element-wise imputation loss and classifier loss, corresponding to the ``imputation" and ``balancing" aspect of IB-GAN respectively.  

\subsubsection{Weighted Resample:} Let $X_{t}$ be a $k$-dimensional multivariate time series with sequence length $m$ with optional metadata $B$ and class labels $Y$. For a given mini-batch size $n_{mb}$, the weighted resampling step for $(X, Y) := (X_{t}, B, Y)$ will first sample real data $X_{real}$ and corresponding labels from the true training data. For synthetic  $(X', Y')$, we sample again $X_{mask}$ (before masking) with weighted probability $\frac{1/w_{y}}{\sum_{Y} 1/w_{y}}$ from each class (denote the corresponding random variable as $Y'$). Together, $X_{real}$ and $X_{mask}$ have balanced samples from each class.

\subsubsection{Data Masking with}$\mathbf{p_{miss}}$:  The proportion of masked values for each sample is specified by hyper-parameter $p_{miss}$. A low $p_{miss}$ ensures imputed samples will be more realistic with greater similarity to original data; a higher $p_{miss}$ encourages greater innovation and variety. It can be seen as a form of regularization on the generator imputation variability. At $p_{miss} = 1$, IB-GAN is analogous to combining a Conditional GAN generating data from white noise with a classifier (Naive GAN); at $p_{miss} = 0$, IB-GAN is equivalent to training on weighted bootstraps of the original data. Optimal choice for $p_{miss}$ will likely depend on class ratio and sample size, and can be found via grid-search in practice. 

Let $\mathbf{I}$ be the component-wise indicator for whether each component of $X_{mask}$ is masked (replaced with random white noise) or keeps the real value. $\mathbf{I}$ has same dimension as the data and the randomly masked data vectors $X_{mask} = X \odot (\boldsymbol{1}-\mathbf{I}) + Z \odot \mathbf{I}$. 


\subsubsection{IB-GAN Triplet Training:} The generator takes in $(X_{mask}, Y')$ as input and impute synthetic samples $(X', Y')$, where $Y'$ are not synthetic labels but rather the true class labels for the imputations, a key difference from Triple-GAN \cite{li2017triple}.  Let the prior label probability of the synthetic data be $w'_{y}=P(Y'=y)$ for $y\in \cal{Y}$ and the conditional probability $p'_{y}(x):=P(X'=x|Y'=y)$ which depends on the generator $G$. Both the true and synthetic samples $(X_{real}, Y)$ and $(X', Y')$ are now inputs to the discriminator and classifier. The classifier learns on a class-balanced sample, and the discriminator attempts to recover $\mathbf{I}$ via the element-wise imputation loss. If the generator is performing well, the discriminator should find it difficult to distinguish between true and imputed values. 

We then jointly optimize the triplet $(D,C,G)$ through combined losses and hyper-parameter $0<\alpha<1$ (Eq. \ref{eq:ibgan_loss}), where we weigh the augmented samples during training of the classifier through a weight that depends on the current discriminator $D(\boldsymbol{x})=\{d_{y}(\boldsymbol{x})\}_{y\in{\cal Y}}$, that is,  $w_{D,y}(x)=\frac{d_{y}(\boldsymbol{x})}{1-d_{y}(\boldsymbol{x})}$. Through these weights, the effect of augmented sample kicks in smoothly during the training of the classifier.

\vspace*{-\abovedisplayskip}
\begin{equation} \label{eq:ibgan_loss}
\begin{aligned}
\min_{G}\max_{C}\max_{D} \{L_{GAN}(D, G) + L_{class}^{\alpha}(C, D, G)\}
\end{aligned}
\end{equation}
\vspace*{-\belowdisplayskip}

\vspace*{-\abovedisplayskip}

\begin{equation}
\resizebox{0.5\textwidth}{!}{%
$\begin{aligned}L_{GAN}(D, G) &= \sum_{y\in{\cal Y}}w_{y}^{-1}\mathbb{E}\left[1\{Y=y\}\log D(\boldsymbol{X})\right] \\ &+\sum_{y\in{\cal Y}}(w'_{y})^{-1} \mathbb{E}\left[1\{Y'=y\}\log\left(1-D(G(X_{mask}))\right)\right]\\
&= \sum_{y\in{\cal Y},\boldsymbol{x}\in{\cal X}^{k}}\left(p_{y}(\boldsymbol{x})\log d_{y}(\boldsymbol{x})+p_{y}'(\boldsymbol{x})\log(1-d_{y}(\boldsymbol{x}))\right)
\end{aligned}$
}
\end{equation}

\vspace*{-\abovedisplayskip}
\begin{equation}
\resizebox{0.49\textwidth}{!}{%
$\begin{aligned}
L_{class}^{\alpha}(C, D, G) &= \alpha\sum_{y\in{\cal Y}}\mathbb{E}\left[1\{Y=y\}\log c_{y}(\boldsymbol{X})\right] \ +  (1-\alpha) \\& \times \sum_{y\in{\cal Y}}\mathbb{E}\left[w_{D,y}1\{Y'=y\}\log c_{y}(G(X_{mask}))\right]\\
= & \sum_{y\in{\cal Y},\boldsymbol{x}\in{\cal X}^{k}} \left[
\alpha w_{y}p_{y}(\boldsymbol{x}) + (1-\alpha) w'_{y}p{}_{y}'(\boldsymbol{x})w_{D,y}(\boldsymbol{x})\log c_{y}(\boldsymbol{x}) \right]
\end{aligned}$%
}
\end{equation}


During initial epochs, the generator will impute lower-quality synthetic samples and the discriminator easily identifies fake vs. real components in $X'$. Hence, the weights $w_{D,y}(x)$ are approximately zero and the classifier is mainly updated through real samples. When the discriminator is close to the optimal $D_{*}$ in the original GAN setting where $D_{*}(\boldsymbol{x})=\{d_{y}^{*}(\boldsymbol{x})\}_{y\in{\cal Y}},d_{y}^{*}(\boldsymbol{x})=\frac{p_{y}(\boldsymbol{x})}{p_{y}(\boldsymbol{x})+p_{y}'(\boldsymbol{x})}$. Then, $p_{y}'(\boldsymbol{x})w_{D,y}(\boldsymbol{x})\approx p_{y}'(\boldsymbol{x})\frac{d_{y}^{*}(\boldsymbol{x})}{1-d_{y}^{*}(\boldsymbol{x})}=p_{y}(\boldsymbol{x})$, and the augmented samples contribute to IB-GAN classifier equivalently to real samples.

\section{Empirical Evaluation}
\subsection{UCR MTS Classification.}\label{subsec:mts_class}
We first apply IB-GAN to open-source multivariate time-series datasets from the popular UCR archive \cite{dau2019ucr}. CharacterTrajectories is a 20-class MTSC dataset with ~3000 samples and Epilepsy is a smaller 4-class MTSC dataset with ~300 samples. Data imbalance is introduced by randomly dropping 75\% of half of the classes.

Two classifier architectures are compared to showcase the flexibility of IB-GAN framework: a simple CNN with 2 Conv-1D layers and a state-of-the-art VGG-like network (3 VGG blocks with Conv2D and max pooling layers). Each classifier choice is paired with 3 GAN architectures: conditional GAN, vanilla GAN, and InfoGAN. Standard baselines are classifier only, class weights, upsampling, and downsampling. State-of-the-art GAN baselines are AC-GAN, BAGAN, SMOTE, and Naive GAN (equivalent to generating data from white noise via Conditional GAN). Filter size for convolution layers is $k$, the time series dimension; $p_{miss} = 0.1$.

\textbf{Findings.} Each experiment is run for 5 replicates, and average classifier performance is listed in Table \ref{summary_perf_UCR}. The two metrics are Balanced Accuracy \cite{kelleher2020fundamentals} and F1-score; both are macro-averaged across classes and penalizes for low performance in minority classes. The default IB-GAN classifier, with Conditional GAN for imputation as the generative model, outperforms against all baselines. IB-VanillaGAN and IB-InfoGAN demonstrate comparable high performance when utilizing the more powerful VGG classifier. 

With both types of classifiers, SOTA GAN baselines such as AC-GAN and BAGAN did not show significant improvements compared to standard baselines, e.g., upsampling. The superior performance of the 3 IB-GAN variants compared to two-stage processes such as AC-GAN and BAGAN testifies that joint training for data augmentation and classification is a far more effective approach. In comparison, the folded classifier in AC-GAN and additional autoencoder-decoder in BAGAN did not generate better quality samples that translated to classification gains. Comparison with Naive GAN also indicates that samples generated via imputation based on randomly masked data vectors (IB-GAN) contributed to a greater performance boost than samples generated via white noise (Naive GAN). Even with a small injection of novelty with $p_{miss}$ at 10\%, the IB-GAN yields significantly higher Balanced Accuracy and F1-score. Finally, the imputation-balancing approach translates to IB-GAN classification with lower variance (see error bars), and stable prediction results are important for downstream applications.

The parametric SMOTE method also has relatively low performance, especially for smaller data size; SMOTE performs poorly for Epilepsy but has comparable performance to BAGAN and AC-GAN for CharacterTrajectories. SMOTE is constrained by forming new samples via linear combinations of existing samples, while IB-GAN functions consistently across sample sizes (Section \ref{subsec: vary_sample_size}). Details for additional UCR experiments and Kaggle EEG time series ($m>1000$) with LSTM classifier are reported in Table \ref{additional_UCR_exp} and \ref{table_brainwave} of Appendix. This testifies to the wide applicability of IB-GAN to time series of different lengths and flexibility of classifier choice.

\begin{figure}[htp]
    \centering
    \includegraphics[width=0.45\textwidth]{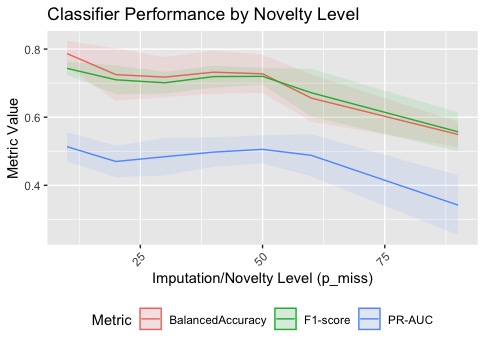}
    \caption{IB-GAN performance as $p_{miss}$ increases, a form of regularization on the generator and level of novelty or innovations in synthetic samples.}
    \label{fig:pmiss_ablation}
\end{figure}

\begin{figure*}[htp]
    \centering
    \includegraphics[width=0.7\textwidth]{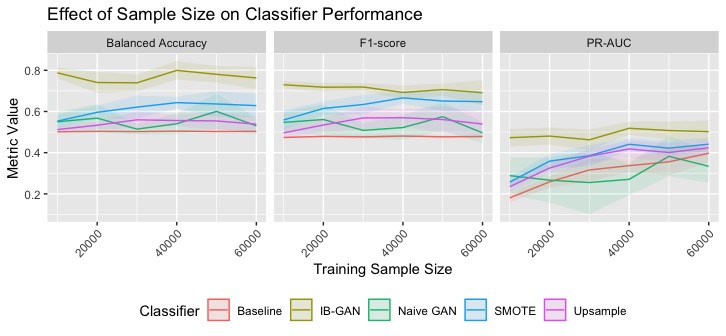}
    \caption{Average Balanced Accuracy, F1-score, and PR-AUC by training sample sizes for Imputation Balanced GAN and baselines evaluated on 30k test set for trending products data.  \label{fig:samp_size}}
\end{figure*}

\begin{table*}[t] \centering 
  \caption{UCR Datasets - Average prediction metrics over 5 replicates for Imputation Balanced GAN and baselines with respect to standard multi-class CNN and VGG classifiers.} 
  \label{summary_perf_UCR} 
  \resizebox{0.85\textwidth}{!}{%
\begin{tabular}{@{\extracolsep{1pt}} lcc|cc} 
\\[-1.8ex]\hline 
\hline \\[-1.8ex] 
& \multicolumn{2}{c}{CNN Classifier: CharacterTrajectories} & \multicolumn{2}{c}{VGG Classifier: CharacterTrajectories} \\ 
\hline \\[-1.8ex] 
Experiment & Balanced Accuracy & F1-score & Balanced Accuracy & F1-score \\ 
\hline \\[-1.8ex] 
IB-GAN & $\mathbf{0.812}$ $\pm$ $0.032$ & $\mathbf{0.810}$ $\pm$ $0.035$ & $\mathbf{0.926}$ $\pm$ $0.009$ & $\mathbf{0.905}$ $\pm$ $0.015$ \\ 
IB-InfoGAN & $0.611$ $\pm$ $0.037$  & $0.535$ $\pm$  $0.043$ & $0.828$ $\pm$ $0.039$ & $0.776$ $\pm$ $0.051$ \\
IB-VanillaGAN & $0.651$ $\pm$  $0.039$  & $0.583$ $\pm$ $0.043$  & $0.862$ $\pm$ $0.039$ & $0.818$ $\pm$ $0.049$\\
\hline
Conditional (Naive) GAN & $0.455$ $\pm$ $0.069$ & $0.416$ $\pm$ $0.075$ & $0.795$ $\pm$ $0.068$ & $0.786$ $\pm$ $0.067$\\
ACGAN + Second Stage Classifier  & $0.313$ $\pm$ $0.075$  & $0.276$ $\pm$ $0.067$ & $0.778$ $\pm$ $0.040$ & $0.772$ $\pm$ $0.033$\\
BAGAN + Second Stage Classifier  & $0.357$ $\pm$  $0.066$  & $0.326$ $\pm$ $0.073$ &  $0.570$ $\pm$ $0.087$ & $0.545$ $\pm$ $0.096$  \\
SMOTE + Second Stage Classifier & $0.314$ $\pm$ $0.095$ & $0.257$ $\pm$ $0.084$ & $0.648$ $\pm$ $0.083$ & $0.595$ $\pm$ $0.090$\\ 
\hline
Baseline Class Weights & $0.332$ $\pm$ $0.024$ & $0.284$ $\pm$ $0.023$ & $0.419$ $\pm$ $0.110$ & $0.344$ $\pm$ $0.107$\\ 
Baseline Upsample & $0.457$ $\pm$ $0.070$ & $0.421$ $\pm$ $0.080$ & $0.683$ $\pm$ $0.095$ & $0.643$ $\pm$ $0.103$\\ 
Baseline Downsample & $0.293$ $\pm$ $0.062$ & $0.246$ $\pm$ $0.069$ & $0.148$ $\pm$ $0.049$ & $0.092$ $\pm$ $0.037$ \\ 
Baseline Classifier & $0.357$ $\pm$ $0.037$ & $0.313$ $\pm$ $0.038$ & $0.393$ $\pm$ $0.061$ & $0.340$ $\pm$ $0.067$\\ 
\hline \\[-1.8ex] 
\end{tabular}%
}
\resizebox{0.85\textwidth}{!}{%
\begin{tabular}{@{\extracolsep{2pt}} lcc|cc} 
\\[-1.8ex]\hline 
\hline \\[-1.8ex] 
& \multicolumn{2}{c}{CNN Classifier: Epilepsy} & \multicolumn{2}{c}{VGG Classifier: Epilepsy} \\ 
\hline \\[-1.8ex] 
Experiment & Balanced Accuracy & F1-score & Balanced Accuracy & F1-score  \\ 
\hline \\[-1.8ex] 
IB-GAN & $\mathbf{0.568}$ $\pm$ $0.055$ & $\mathbf{0.549}$ $\pm$ $0.045$ & $\mathbf{0.767}$ $\pm$ $0.080$ & $\mathbf{0.694}$ $\pm$ $0.111$ \\ 
IB-InfoGAN   &  $0.428$ $\pm$ $0.062$ & $0.381$ $\pm$ $0.072$ & $0.746$ $\pm$ $0.037$ & $0.681$ $\pm$ $0.043$\\ 
IB-VanillaGAN  &  $0.339$ $\pm$ $0.098$ & $0.308$ $\pm$ $0.088$  & $0.722$ $\pm$ $0.046$ & $0.652$ $\pm$ $0.062$\\ 
\hline
Conditional (Naive) GAN & $0.343$ $\pm$ $0.105$ & $0.305$ $\pm$ $0.100$ & $0.487$ $\pm$ $0.147$ & $0.456$ $\pm$ $0.154$\\
ACGAN + Second Stage Classifier  & $0.318$ $\pm$ $0.102$ & $0.295$ $\pm$ $0.095$  &  $0.441$ $\pm$ $0.176$ & $0.352$ $\pm$  $0.190$ \\
BAGAN + Second Stage Classifier  &  $0.224$ $\pm$ $0.055$ & $0.205$ $\pm$ $0.052$ &  $0.280$ $\pm$ $0.066$ & $0.158$ $\pm$ $0.103$ \\
SMOTE + Second Stage Classifier & $0.248$ $\pm$ $0.029$ & $0.227$ $\pm$ $0.030$ & $0.275$ $\pm$ $0.035$ & $0.160$ $\pm$ $0.059$\\ 
 \hline \\[-1.8ex] 
Baseline Class Weights & $0.307$ $\pm$ $0.033$ & $0.276$ $\pm$ $0.038$ & $0.284$ $\pm$ $0.025$ & $0.192$ $\pm$ $0.043$\\ 
Baseline Upsample & $0.250$ $\pm$ $0.043$ & $0.235$ $\pm$ $0.056$ & $0.330$ $\pm$ $0.083$ & $0.222$ $\pm$ $0.113$\\ 
Baseline Downsample & $0.266$ $\pm$ $0.018$ & $0.219$ $\pm$ $0.053$ & $0.260$ $\pm$ $0.060$ & $0.195$ $\pm$ $0.065$ \\
Baseline Classifier & $0.287$ $\pm$ $0.039$ & $0.237$ $\pm$ $0.058$ & $0.270$ $\pm$ $0.034$ & $0.156$ $\pm$ $0.037$ \\ 
\hline \\[-1.8ex] 
\end{tabular}%
}
\end{table*}
\vspace*{-\belowdisplayskip}

\subsection{Predicting Trending Products.}
We now apply IB-GAN to an empirical problem setting. The motivating application is predicting trending products on a large e-commerce website based on past time series metrics, which make up a minority class of all newly-launched products. This is known as the \textit{cold-start} problem in search, learning-to-rank, and recommendation systems. An additional challenge is the existence of time-invariant features based on product metadata, which will correspond to $B$ in the IB-GAN set-up.

We can frame this problem as a large-scale imbalanced MTSC task with a proprietary 90K dataset of products, split into 60K training and 30K testing. Each item has a binary label $Y$ which indicates it is a top-ranked item, 15 time-invariant metadata features $B$ and 8 daily time series $X_{t}$ for 3 weeks. The full feature set is the joint vector $X = \{B, X_{t}\}$. B is different from class labels Y; it is additional auxiliary information, e.g., product characteristics. Class imbalance is strong with only 11\% samples in the 1-class. Standard accuracy is an inappropriate measure; an initial baseline CNN classifier with 89\% testing accuracy has a precision of 0.29 and recall of 0.01 for the 1-class. 

Given this large dataset, we conduct ablation experiments for IB-GAN training: (1) \textit{How does choice of $p_{miss}$ affect IB-GAN generator quality and classification accuracy?} (2) \textit{How does IB-GAN perform across  different training sample sizes?}

\subsubsection{Novelty Parameter Tuning.}
Using the full product dataset, we compare IB-GAN classifier performance at different $p_{miss}$ levels against benchmarks. We use a state-of-the-art VGG classifier with blocks of Conv1D, batch normalization, and max pooling layers; a similar architecture is adopted for the discriminator-generator duo, which conditions on class labels analogous to Conditional GAN. Each model is run for 100 epochs across 10 replicates. Figure \ref{fig:pmiss_ablation} shows that 10\% is a good default value, with an average Balanced Accuracy of 78.8\%. The nearest benchmark is SMOTE with 64.1\%. The three metrics are fairly consistent when $p_{miss}$ is between 20-45\%. As novelty in synthetic samples increase past 40-50\%, classification performance declines on average. Higher $p_{miss}$ dictates greater variability in generator output, which results in synthetic samples that are less similar to original data. This leads to greater variability in prediction accuracy; see error bands for F1-score and PR-AUC. See Table ~\ref{table_60K} in Appendix for results of standard and GAN baselines.

\subsubsection{Training Data Size.} \label{subsec: vary_sample_size}

To understand IB-GAN performance across sample sizes, we fix $p_{miss} = 20\%$ and randomly sample subsets of the 60K training data. The same VGG classifier and IB-GAN set-up is used as the previous experiment. Figure \ref{fig:samp_size} shows average Balanced Accuracy, F1-score, and PR-AUC over 10 replicates as evaluated on the fixed 30K test set for IB-GAN and select baselines.

The IB-GAN classifier outperforms all benchmarks with consistent Balanced Accuracy and F1-score across sample sizes up to 60000, the maximum number of training samples. As sample size increases, upsampling and SMOTE become more effective due to increased examples from the minority class. When sample sizes are small, upsampling leads to overfitting and SMOTE is unable to interpolate with variety to create meaningful new training samples. Meanwhile, GAN-based methods such as IB-GAN and even Naive GAN largely avoids this issue, and are recommended for applications with small data sizes. Error bars for IB-GAN metrics tend to be narrower than Naive GAN; the imputation mechanism and tuning parameter $p_{miss}$ both regulate IB-GAN generator variability and ensure synthetic samples generated actually enables classifier to learn signals from minority classes. See Table \ref{model_perf_samplesize} for details.



\section{Significance and Impact}

We have proposed a novel IB-GAN method for joint data augmentation and multivariate time series classification with highly imbalanced data in an unified one-step process. The framework consists of a triplet of generator $G$, discriminator $D$, and classifier $C$ models that works seamlessly with choices of GAN architectures and deep learning classifiers. Compared to prior methods, IB-GAN does not require a cumbersome second stage classifier, and directly incorporates classification loss as feedback to improve synthetic data quality. IB-GAN uses a unique imputation and balancing approach that leverages existing training data to generate higher quality synthetic samples, and enables better classification performance on under-observed classes by learning on a class-balanced training set. The hyperparameter $p_{miss}$ directly tunes the similarity vs. novelty in synthetic samples while side-stepping mode collapse from standard GAN training. Empirical results from UCR datasets show IB-GAN and its variants achieving significant performance against state-of-the-art parametric and GAN baselines. Ablation studies with a trending product dataset demonstrate IB-GAN performance is robust across sample sizes (narrower confidence intervals) and $p_{miss}$ levels up to 50\% under MCAR.

The IB-GAN framework is quite versatile and can easily be extended to more complex datasets such as images, video, and text; future work can identify the proper design of generator-discriminator and classifier architecture suitable for these tasks. IB-GAN can also be applied to mitigate data bias and fairness issues in many ML applications; see Section \ref{subsec:fairness} in Appendix for a short note on fairness.



\bibliography{ref}

\newpage
\onecolumn

\section{Appendix: A Short Note on Fairness} \label{subsec:fairness}
The IB-GAN can potentially be extended to mitigate data bias in algorithmic and AI fairness \cite{dwork2012fairness, corbett2017algorithmic, mehrabi2019survey}. A key source of data bias \cite{dressel2018accuracy, olteanu2019social} is class imbalance for minority or protected classes (gender, race, age, etc.). Imbalanced classification leads to poor performance in both standard metrics and fairness metrics such as statistical parity, demographic parity or predictive parity. Prior work considered data augmentation techniques (e.g., SMOTE) to generate synthetic samples for minority classes and improve model fairness for financial credit, education, and other social policies \cite{iosifidis2019adafair, hutt2019evaluating}. IB-GAN with the optimal generator $G^{*}$ also imputes synthetic samples with an emphasis on minority classes of interest, such that $X' \sim p(X)$, the true data distribution. The IB-GAN classifier learns from a balanced dataset with equal representation from each class, and outperforms SMOTE and other data augmentation techniques for under-observed classes. Given datasets in this work contain no sensitive attributes, the authors consider improvements in specific fairness metrics via IB-GAN as an open problem and a future research direction.

\section{Appendix: Additional Experiment Details} \label{sec: appendixB}

\subsection{Predicting Trending Products.}
Table \ref{table_60K} reports macro-averaged Balanced Accuracy (equivalent to macro-averaged Recall in \texttt{scikit-learn}), F1-score, and PR-AUC with standard errors evaluated on a fixed 30K test product dataset, for 10 replicates of IB-GAN classifiers with different $p_{miss}$ levels. Table \ref{model_perf_samplesize} reports the average Balanced Accuracy, F1-score, and PR-AUC with standard errors, also evaluated on the test dataset, for 10 replicates of IB-GAN classifiers with a fixed $p_{miss} = 0.2$ and various training sample sizes. 

Performance for various baselines are also reported. Naive GAN is analagous to IB-GAN with 100\% missingness ($p_{miss}$ = 1), where the generator is initialized with white noise. The IB-GAN generator and discriminator follow a Conditional GAN architecture where the multivariate time series metrics $X_{t}$, metadata $B$, and class labels $Y$ are all taken as inputs. Each replicate takes about 30 minutes on a ml.m5.24xlarge instance.

\begin{table*}[htp] \centering 
  \caption{Trending Products: Classifier Performance by Imputation Level, 10 Replicates. \label{table_60K}} 
 \resizebox{0.65\textwidth}{!}{%
\begin{tabular}{@{\extracolsep{1pt}} ccccc} 
\\[-1.8ex]\hline 
\hline \\[-1.8ex] 
Classifier Type & Balanced Accuracy  & F1-score  & PR-AUC\\ 
\hline \\[-1.8ex] 
10\% Imputed IB-GAN & $0.787$ $\pm$ $0.038$ & $0.743$ $\pm$ $0.019$ & $0.513$ $\pm$ $0.043$ \\ 
20\% Imputed IB-GAN & $0.725$ $\pm$ $0.077$ & $0.710$ $\pm$ $0.042$ & $0.470$ $\pm$ $0.046$ \\ 
30\% Imputed IB-GAN & $0.718$ $\pm$ $0.059$ & $0.701$ $\pm$ $0.032$ & $0.484$ $\pm$ $0.055$ \\ 
40\% Imputed IB-GAN & $0.732$ $\pm$ $0.064$ & $0.719$ $\pm$ $0.032$ & $0.498$ $\pm$ $0.043$ \\ 
50\% Imputed IB-GAN & $0.727$ $\pm$ $0.057$ & $0.720$ $\pm$ $0.025$ & $0.506$ $\pm$ $0.041$ \\ 
\hline \\[-1.8ex] 
Naive GAN & $0.549$ $\pm$ $0.038$ & $0.557$ $\pm$ $0.057$ & $0.342$ $\pm$ $0.088$ \\ 
SMOTE & $0.641$ $\pm$ $0.059$ & $0.660$ $\pm$ $0.047$ & $0.440$ $\pm$ $0.064$ \\ 
Baseline & $0.504$ $\pm$ $0.005$ & $0.480$ $\pm$ $0.012$ & $0.363$ $\pm$ $0.046$ \\ 
Baseline Class Weights & $0.646$ $\pm$ $0.058$ & $0.648$ $\pm$ $0.031$ & $0.364$ $\pm$ $0.046$ \\ 
Baseline Downsample & $0.635$ $\pm$ $0.041$ & $0.568$ $\pm$ $0.052$ & $0.248$ $\pm$ $0.054$ \\ 
Metadata Only & $0.584$ $\pm$ $0.016$ & $0.528$ $\pm$ $0.027$ & $0.220$ $\pm$ $0.012$ \\ 
Baseline Upsample & $0.560$ $\pm$ $0.035$ & $0.575$ $\pm$ $0.049$ & $0.432$ $\pm$ $0.048$ \\ 
\hline \\[-1.8ex] 
\end{tabular}%
}
\end{table*} 
\vspace*{-\belowdisplayskip}

\begin{table}[htp] \centering 
  \caption{Trending Products: Classifier Performance by Sample Size, 10 Replicates.} 
  \label{model_perf_samplesize} 
\begin{tabular}{@{\extracolsep{1pt}} c|l|cccc} 
\\[-1.8ex]\hline 
\hline \\[-1.8ex] 
Sample Size & Classifier Type & Balanced Accuracy & F1-score & PR-AUC \\ 
\hline \\[-1.8ex] 
$10,000$ & IB-GAN & $0.788$ $\pm$ $0.027$ & $0.730$ $\pm$ $0.018$ & $0.473$ $\pm$ $0.043$ \\ 
 & Naive GAN & $0.550$ $\pm$ $0.049$ & $0.547$ $\pm$ $0.052$ & $0.289$ $\pm$ $0.089$ \\
 & SMOTE & $0.553$ $\pm$ $0.033$ & $0.559$ $\pm$ $0.047$ & $0.257$ $\pm$ $0.026$ \\
& Baseline Class Weights & $0.581$ $\pm$ $0.038$ & $0.524$ $\pm$ $0.056$ & $0.210$ $\pm$ $0.037$ \\ 
 & Baseline Downsample & $0.540$ $\pm$ $0.034$ & $0.426$ $\pm$ $0.106$ & $0.158$ $\pm$ $0.033$ \\ 
 & Baseline  & $0.501$ $\pm$ $0.002$ & $0.474$ $\pm$ $0.004$ & $0.181$ $\pm$ $0.024$ \\ 
 & Baseline Upsample & $0.512$ $\pm$ $0.013$ & $0.496$ $\pm$ $0.027$ & $0.235$ $\pm$ $0.028$ \\ 
\hline \\[-1.8ex] 
$20,000$ & IB-GAN & $0.741$ $\pm$ $0.052$ & $0.718$ $\pm$ $0.023$ & $0.481$ $\pm$ $0.042$ \\ 
 & Naive GAN & $0.568$ $\pm$ $0.070$ & $0.561$ $\pm$ $0.073$ & $0.267$ $\pm$ $0.111$ \\
 & SMOTE & $0.596$ $\pm$ $0.033$ & $0.615$ $\pm$ $0.034$ & $0.359$ $\pm$ $0.034$ \\ 
 & Baseline Class Weights & $0.623$ $\pm$ $0.024$ & $0.588$ $\pm$ $0.055$ & $0.260$ $\pm$ $0.027$ \\ 
 & Baseline Downsample & $0.562$ $\pm$ $0.029$ & $0.464$ $\pm$ $0.067$ & $0.175$ $\pm$ $0.036$ \\ 
 & Baseline  & $0.503$ $\pm$ $0.002$ & $0.479$ $\pm$ $0.005$ & $0.258$ $\pm$ $0.028$ \\ 
 & Baseline Upsample & $0.534$ $\pm$ $0.020$ & $0.534$ $\pm$ $0.034$ & $0.325$ $\pm$ $0.028$ \\ 
\hline \\[-1.8ex] 
$30,000$ & IB-GAN & $0.739$ $\pm$ $0.042$ & $0.719$ $\pm$ $0.019$ & $0.463$ $\pm$ $0.051$ \\ 
 & Naive GAN & $0.514$ $\pm$ $0.025$ & $0.508$ $\pm$ $0.034$ & $0.255$ $\pm$ $0.154$ \\ 
 & SMOTE & $0.621$ $\pm$ $0.057$ & $0.634$ $\pm$ $0.047$ & $0.385$ $\pm$ $0.052$ \\ 
 & Baseline Class Weights & $0.652$ $\pm$ $0.061$ & $0.639$ $\pm$ $0.041$ & $0.319$ $\pm$ $0.058$ \\ 
 & Baseline Downsample & $0.602$ $\pm$ $0.033$ & $0.511$ $\pm$ $0.048$ & $0.206$ $\pm$ $0.030$ \\ 
 & Baseline  & $0.502$ $\pm$ $0.003$ & $0.477$ $\pm$ $0.007$ & $0.316$ $\pm$ $0.054$ \\ 
 & Baseline Upsample & $0.560$ $\pm$ $0.040$ & $0.569$ $\pm$ $0.054$ & $0.383$ $\pm$ $0.035$ \\ 
\hline \\[-1.8ex] 
$40,000$ & IB-GAN & $0.800$ $\pm$ $0.045$ & $0.692$ $\pm$ $0.035$ & $0.519$ $\pm$ $0.033$ \\ 
 & Naive GAN & $0.541$ $\pm$ $0.051$ & $0.522$ $\pm$ $0.060$ & $0.271$ $\pm$ $0.077$ \\ 
 & SMOTE & $0.643$ $\pm$ $0.029$ & $0.666$ $\pm$ $0.028$ & $0.441$ $\pm$ $0.029$ \\
 & Baseline Class Weights & $0.670$ $\pm$ $0.055$ & $0.659$ $\pm$ $0.033$ & $0.342$ $\pm$ $0.044$ \\ 
 & Baseline Downsample & $0.597$ $\pm$ $0.029$ & $0.525$ $\pm$ $0.041$ & $0.201$ $\pm$ $0.016$ \\ 
 & Baseline  & $0.504$ $\pm$ $0.003$ & $0.481$ $\pm$ $0.007$ & $0.337$ $\pm$ $0.033$ \\ 
 & Baseline Upsample & $0.556$ $\pm$ $0.028$ & $0.570$ $\pm$ $0.041$ & $0.418$ $\pm$ $0.041$ \\ 
\hline \\[-1.8ex] 
$50,000$ & IB-GAN & $0.780$ $\pm$ $0.040$ & $0.706$ $\pm$ $0.030$ & $0.508$ $\pm$ $0.045$ \\ 
 & Naive GAN & $0.601$ $\pm$ $0.086$ & $0.575$ $\pm$ $0.069$ & $0.382$ $\pm$ $0.097$ \\ 
 & SMOTE & $0.637$ $\pm$ $0.063$ & $0.651$ $\pm$ $0.047$ & $0.422$ $\pm$ $0.036$ \\ 
 & Baseline Class Weights & $0.644$ $\pm$ $0.050$ & $0.649$ $\pm$ $0.035$ & $0.360$ $\pm$ $0.027$ \\ 
 & Baseline Downsample & $0.607$ $\pm$ $0.024$ & $0.574$ $\pm$ $0.042$ & $0.235$ $\pm$ $0.023$ \\ 
 & Baseline  & $0.503$ $\pm$ $0.003$ & $0.477$ $\pm$ $0.006$ & $0.356$ $\pm$ $0.065$ \\ 
 & Baseline Upsample & $0.554$ $\pm$ $0.043$ & $0.561$ $\pm$ $0.065$ & $0.401$ $\pm$ $0.047$ \\ 
\hline \\[-1.8ex] 
$60,000$ & IB-GAN & $0.763$ $\pm$ $0.054$ & $0.691$ $\pm$ $0.061$ & $0.502$ $\pm$ $0.054$ \\ 
 & Naive GAN & $0.530$ $\pm$ $0.041$ & $0.496$ $\pm$ $0.052$ & $0.334$ $\pm$ $0.080$ \\
 & SMOTE & $0.629$ $\pm$ $0.057$ & $0.648$ $\pm$ $0.046$ & $0.441$ $\pm$ $0.032$ \\ 
 & Baseline Class Weights & $0.677$ $\pm$ $0.064$ & $0.665$ $\pm$ $0.033$ & $0.383$ $\pm$ $0.040$ \\ 
 & Baseline Downsample & $0.631$ $\pm$ $0.029$ & $0.600$ $\pm$ $0.023$ & $0.263$ $\pm$ $0.028$ \\ 
 & Baseline  & $0.503$ $\pm$ $0.003$ & $0.479$ $\pm$ $0.008$ & $0.398$ $\pm$ $0.044$ \\ 
 & Baseline Upsample & $0.538$ $\pm$ $0.034$ & $0.539$ $\pm$ $0.057$ & $0.424$ $\pm$ $0.036$ \\ 
\hline \\[-1.8ex] 
\end{tabular} 
\end{table}

\subsection{Multivariate Time Series Classification with UCR Data.}
Table \ref{summary_perf_UCR} reports macro-averaged Balanced Accuracy, F1-score, and PR-AUC with standard errors for IB-GAN classifier with a fixed $p_{miss} = 0.1$ against select benchmarks for 8 additional open-source MTS datasets in the UCR archive. The data has already been pre-processed and split into training and testing datasets. We utilize the same VGG architecture for all GAN and baseline classifiers, with 2 Conv1D layers each with relu-activation, max pooling and batch normalization. For IB-GAN and Naive GAN, the discriminator and generator models also have 2 Conv1D layers with relu-activation.  Each replicate trains for 20 epochs and takes about 20 minutes to train.

\begin{table}[htp] \centering 
  \caption{UCR MTS Datasets: Classifier Performance, 5 Replicates.} 
  \label{additional_UCR_exp} 
  \resizebox{0.7\textwidth}{!}{%
\begin{tabular}{@{\extracolsep{1pt}} c|l|cc} 
\\[-1.8ex]\hline 
\hline \\[-1.8ex] 
Dataset & Classifier Type & Balanced Accuracy & F1-score \\ 
\hline \\[-1.8ex] 
FingerMovements & IB-GAN & $0.529$ $\pm$ $0.052$ & $0.482$ $\pm$ $0.104$ \\ 
 & Naive GAN & $0.477$ $\pm$ $0.081$ & $0.467$ $\pm$ $0.081$ \\
 & SMOTE & $0.483$ $\pm$ $0.054$ & $0.369$ $\pm$ $0.116$ \\ 
 & Baseline Weights & $0.494$ $\pm$ $0.042$ & $0.468$ $\pm$ $0.060$ \\ 
 & Baseline Upsample & $0.489$ $\pm$ $0.057$ & $0.389$ $\pm$ $0.135$ \\ 
 & Baseline  & $0.487$ $\pm$ $0.094$ & $0.424$ $\pm$ $0.082$ \\ 
 & Baseline Downsample & $0.500$ $\pm$ $0.034$ & $0.393$ $\pm$ $0.113$ \\ 
HandMovementDirection & IB-GAN & $0.422$ $\pm$ $0.108$ & $0.345$ $\pm$ $0.094$ \\ 
& Naive GAN & $0.258$ $\pm$ $0.078$ & $0.238$ $\pm$ $0.056$ \\
& SMOTE & $0.261$ $\pm$ $0.063$ & $0.225$ $\pm$ $0.039$ \\ 
& Baseline Weights & $0.204$ $\pm$ $0.058$ & $0.194$ $\pm$ $0.033$ \\ 
& Baseline Upsample & $0.236$ $\pm$ $0.095$ & $0.178$ $\pm$ $0.062$ \\
& Baseline  & $0.319$ $\pm$ $0.102$ & $0.237$ $\pm$ $0.045$ \\ 
& Baseline Downsample & $0.291$ $\pm$ $0.114$ & $0.232$ $\pm$ $0.078$ \\ 
Handwriting & IB-GAN & $0.187$ $\pm$ $0.006$ & $0.159$ $\pm$ $0.009$ \\ 
 & Naive GAN & $0.106$ $\pm$ $0.026$ & $0.089$ $\pm$ $0.018$ \\ 
 & SMOTE & $0.067$ $\pm$ $0.026$ & $0.051$ $\pm$ $0.022$ \\ 
 & Baseline Weights & $0.067$ $\pm$ $0.013$ & $0.055$ $\pm$ $0.010$ \\ 
 & Baseline Upsample & $0.049$ $\pm$ $0.018$ & $0.038$ $\pm$ $0.010$ \\ 
 & Baseline  & $0.050$ $\pm$ $0.019$ & $0.043$ $\pm$ $0.018$ \\ 
 & Baseline Downsample & $0.037$ $\pm$ $0.016$ & $0.035$ $\pm$ $0.017$ \\ 
Heartbeat & IB-GAN & $0.571$ $\pm$ $0.043$ & $0.556$ $\pm$ $0.060$ \\ 
 & Naive GAN & $0.514$ $\pm$ $0.016$ & $0.473$ $\pm$ $0.046$ \\
 & SMOTE & $0.498$ $\pm$ $0.024$ & $0.417$ $\pm$ $0.066$ \\ 
 & Baseline Weights & $0.517$ $\pm$ $0.033$ & $0.378$ $\pm$ $0.092$ \\ 
 & Baseline Upsample & $0.502$ $\pm$ $0.024$ & $0.418$ $\pm$ $0.072$ \\ 
 & Baseline  & $0.504$ $\pm$ $0.048$ & $0.450$ $\pm$ $0.070$ \\ 
 & Baseline Downsample & $0.498$ $\pm$ $0.027$ & $0.315$ $\pm$ $0.096$ \\ 
Libras & IB-GAN & $0.191$ $\pm$ $0.079$ & $0.102$ $\pm$ $0.075$ \\ 
 & Naive GAN & $0.123$ $\pm$ $0.044$ & $0.063$ $\pm$ $0.030$ \\ 
 & SMOTE & $0.070$ $\pm$ $0.019$ & $0.031$ $\pm$ $0.020$ \\ 
 & Baseline Weights & $0.080$ $\pm$ $0.021$ & $0.035$ $\pm$ $0.037$ \\
 & Baseline Upsample & $0.050$ $\pm$ $0.022$ & $0.016$ $\pm$ $0.012$ \\ 
 & Baseline  & $0.064$ $\pm$ $0.032$ & $0.033$ $\pm$ $0.030$ \\ 
 & Baseline Downsample & $0.090$ $\pm$ $0.025$ & $0.036$ $\pm$ $0.022$ \\ 
RacketSports & IB-GAN & $0.424$ $\pm$ $0.055$ & $0.367$ $\pm$ $0.075$ \\ 
 & Naive GAN & $0.328$ $\pm$ $0.099$ & $0.296$ $\pm$ $0.081$ \\ 
 & SMOTE & $0.255$ $\pm$ $0.068$ & $0.225$ $\pm$ $0.081$ \\ 
 & Baseline Weights & $0.257$ $\pm$ $0.036$ & $0.206$ $\pm$ $0.039$ \\ 
 & Baseline Upsample & $0.264$ $\pm$ $0.089$ & $0.233$ $\pm$ $0.078$ \\ 
 & Baseline  & $0.222$ $\pm$ $0.041$ & $0.205$ $\pm$ $0.044$ \\ 
 & Baseline Downsample & $0.261$ $\pm$ $0.060$ & $0.209$ $\pm$ $0.055$ \\ 
SelfRegulationSCP1 & IB-GAN & $0.814$ $\pm$ $0.112$ & $0.808$ $\pm$ $0.096$ \\ 
 & Naive GAN & $0.475$ $\pm$ $0.216$ & $0.495$ $\pm$ $0.221$ \\ 
 & SMOTE & $0.744$ $\pm$ $0.077$ & $0.713$ $\pm$ $0.071$ \\ 
 & Baseline Weights & $0.767$ $\pm$ $0.076$ & $0.713$ $\pm$ $0.067$ \\ 
 & Baseline Upsample & $0.706$ $\pm$ $0.114$ & $0.631$ $\pm$ $0.101$ \\ 
 & Baseline  & $0.632$ $\pm$ $0.140$ & $0.590$ $\pm$ $0.114$ \\ 
 & Baseline Downsample & $0.749$ $\pm$ $0.190$ & $0.653$ $\pm$ $0.168$ \\ 
SpokenArabicDigits & IB-GAN & $0.681$ $\pm$ $0.026$ & $0.642$ $\pm$ $0.031$ \\ 
 & Naive GAN & $0.428$ $\pm$ $0.069$ & $0.404$ $\pm$ $0.086$ \\ 
 & SMOTE & $0.622$ $\pm$ $0.056$ & $0.606$ $\pm$ $0.067$ \\ 
 & Baseline Weights & $0.456$ $\pm$ $0.037$ & $0.429$ $\pm$ $0.054$ \\ 
 & Baseline Upsample & $0.627$ $\pm$ $0.053$ & $0.627$ $\pm$ $0.058$ \\ 
 & Baseline  & $0.495$ $\pm$ $0.082$ & $0.476$ $\pm$ $0.093$ \\ 
 & Baseline Downsample & $0.456$ $\pm$ $0.094$ & $0.380$ $\pm$ $0.072$ \\ 
\hline \\[-1.8ex] 
\end{tabular}%
}
\end{table}

\subsection{EEG-Emotion Classification with LSTM.}
To replicate the IB-GAN's success on a much longer sequence and with the popular LSTM classifier for time series, we utilize an open-source Kaggle dataset of EEG brainwave time series to classify the subject's feelings based on EEG: Neutral, Positive, Negative. There are a total of 2132 samples over 2549 time periods, and classes are roughly equally distributed. 50\% of the Neutral class is randomly removed to introduce class imbalance.

\begin{figure}[htp]
    \centering
    \includegraphics[width=0.7\textwidth]{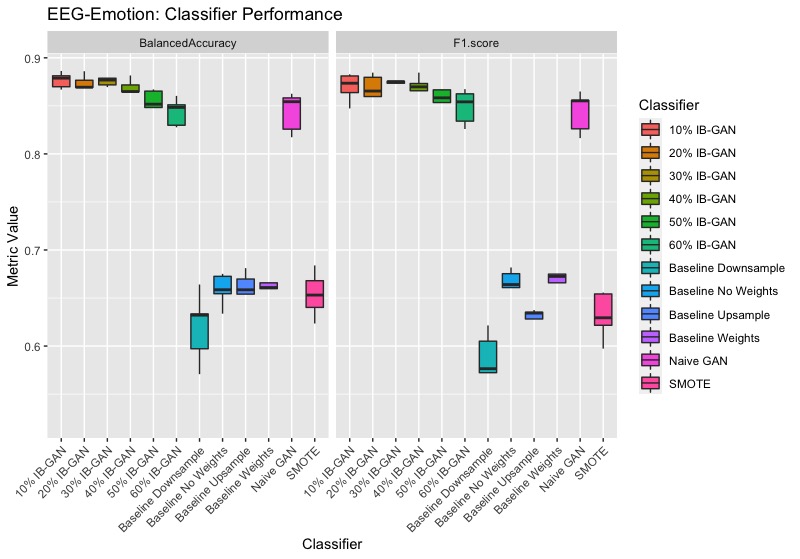}
    \caption{EEG-Emotion Classification with LSTM - Performance at various imputation levels. \label{fig:brainwave-perf}}
\end{figure}

Given the long time series, the IB-GAN classifier is a standard LSTM model with 50 hidden neurons, with a Conditional GAN-like generator-discriminator duo with dense layers. The discriminator uses sigmoid activation to predict the final probability of real or fake sample with 0.5 as the cutoff threshold. The generator uses softmax function to predict the class of each sample. Loss functions for generator and discriminator are binary and categorical cross-entropy respectively. We use the Adam optimizer with default parameters instead of stochastic gradient descent for faster convergence. The hyper-parameter $p_{miss}$ is varied from 10\% to 60\% missing, and classification metrics are compared standard and GAN baselines. The IB-GAN classifier at any level of imputation outperforms all benchmarks in terms of Balanced Accuracy and F1-score, averaging 0.876 Balanced Accuracy at 30\% followed by 0.873 at 10\% and 0.865 at 20\%.The boxplot in Fig.~\ref{fig:brainwave-perf} shows the distribution of the 2 measures across 5 replicates. Table \ref{table_brainwave} reports macro-averaged Balanced Accuracy, F1-score, and PR-AUC with standard errors for IB-GAN classifier at various $p_{miss}$ levels. Each GAN experiment replicate trains for 100 epochs, about 70 minutes.

\vspace*{-\belowdisplayskip}
\vspace*{-\abovedisplayskip}
\begin{table}[htp] \centering 
  \caption{EEG-Emotion: Classifier Performance, 5 Replicates. \label{table_brainwave}} 
 \resizebox{0.55\textwidth}{!}{%
\begin{tabular}{@{\extracolsep{1pt}} ccc} 
\\[-1.8ex]\hline 
\hline \\[-1.8ex] 
Classifier Type & Balanced Accuracy  & F1-score  \\ 
\hline \\[-1.8ex] 
10\% IB-GAN & $0.873$ $\pm$ $0.016$ & $0.870$ $\pm$ $0.014$ \\ 
20\% IB-GAN & $0.865$ $\pm$ $0.025$ & $0.855$ $\pm$ $0.039$ \\ 
30\% IB-GAN & $0.875$ $\pm$ $0.004$ & $0.876$ $\pm$ $0.003$ \\ 
40\% IB-GAN & $0.860$ $\pm$ $0.023$ & $0.866$ $\pm$ $0.018$ \\ 
50\% IB-GAN & $0.851$ $\pm$ $0.019$ & $0.855$ $\pm$ $0.015$ \\ 
60\% IB-GAN & $0.843$ $\pm$ $0.014$ & $0.849$ $\pm$ $0.018$ \\ 
\hline \\[-1.8ex] 
Naive GAN & $0.811$ $\pm$ $0.095$ & $0.812$ $\pm$ $0.094$ \\ 
LSTM SMOTE & $0.654$ $\pm$ $0.024$ & $0.632$ $\pm$ $0.024$ \\ 
LSTM Downsample & $0.620$ $\pm$ $0.036$ & $0.580$ $\pm$ $0.038$ \\ 
LSTM Baseline & $0.659$ $\pm$ $0.017$ & $0.664$ $\pm$ $0.016$ \\ 
LSTM Upsample & $0.654$ $\pm$ $0.028$ & $0.625$ $\pm$ $0.019$ \\ 
LSTM Class Weights & $0.661$ $\pm$ $0.018$ & $0.672$ $\pm$ $0.017$ \\ 
\hline \\[-1.8ex] 
\end{tabular}%
}
\end{table} 
\vspace*{-\belowdisplayskip}

\end{document}